%% file: manuscript.tex
\documentclass[runningheads]{llncs} 
\usepackage{graphicx}
\usepackage{comment}
\usepackage{amsmath}
\usepackage{xcolor}
\PassOptionsToPackage{hyphens}{url} 
\usepackage{hyperref}
\usepackage{cleveref}
\usepackage[normalem]{ulem}
\usepackage{bbm}
\usepackage{marvosym}
\usepackage{float}
\usepackage{multirow}
\usepackage{tabularx}
\usepackage{algorithm} 
\usepackage[noend]{algorithmic}

\usepackage{lipsum}
\crefformat{footnote}{#2\footnotemark[#1]#3}

\newcommand{\repeatthanks}{\textsuperscript{\thefootnote}}

\begin{document}

            Paper accepted at CLAR 2021, Fourth International Conference on Logic and Argumentation, Hangzhou, China, October 20–22, 2021.
            
            \vspace{1cm}
            The final authenticated publication is available online at\\ \url{https://doi.org/10.1007/978-3-030-89391-0_27}.
            
            \vspace{1cm}
            Please cite this work as:\\
            Fazzinga B., Galassi A., Torroni P. (2021) An Argumentative Dialogue System for COVID-19 Vaccine Information. In: Baroni P., Benzmüller C., Wáng Y.N. (eds) Logic and Argumentation. CLAR 2021. Lecture Notes in Computer Science, vol 13040. Springer, Cham. https://doi.org/10.1007/978-3-030-89391-0\_27

%
\title{An Argumentative Dialogue System for COVID-19 Vaccine Information}
%
%
\author{Bettina Fazzinga\thanks{Equal contribution.}\inst{1}\orcidID{0000-0001-8611-2377} \and
Andrea Galassi\repeatthanks\inst{2}\textsuperscript{(\Letter)}\orcidID{0000-0001-9711-7042} \and
Paolo Torroni\repeatthanks\inst{2}\orcidID{0000-0002-9253-8638}}
\authorrunning{B. Fazzinga et al.} 
%
\institute{ICAR-CNR, Rende, Italy\\
 DICES, University of Calabria, Rende, Italy \\
\email{bettina.fazzinga@icar.cnr.it}\\
\and
DISI, University of Bologna, Bologna, Italy\\
\email{\{a.galassi@unibo.it,paolo.torroni\}@unibo.it}}
\maketitle              
\begin{abstract}
Dialogue systems are widely used in AI to support timely and interactive communication with users.
We propose a general-purpose dialogue system architecture that leverages computational argumentation to perform reasoning and provide consistent and explainable answers. We illustrate the system using a COVID-19 vaccine information case study.
\end{abstract}

\section{Introduction}
\label{sec:intro}
\input{s_intro}

\section{Related Work}
\label{sec:related}
\input{s_related}
\section{System Architecture}
\label{sec:argumentation}
\label{sec:architecture}
\input{s_architecture}

\section{Argumentation Module}
\label{sec:framework}
\input{s_frameworkNEW}

\section{Case Study}
\label{sec:case}
\input{s_casestudy}


\section{Conclusion}
\label{sec:conclusions}
\input{s_conclusion}

\section*{Acknowledgments}
The research reported in this work was partially supported by the EU H2020 ICT48 project ``Humane AI Net" under contract \#952026.

%
%
%
%
\bibliographystyle{splncs04}
\bibliography{biblio}

\end{document}

%% file: s_intro.tex


Since the early days of AI, research has been inspired by the idea of developing programs that can communicate with users in natural language.
With the advent of language technologies able to reach human performance in various tasks, AI chatbots and dialogue systems are starting to mature and this vision seems nearer than ever.
As a result, more organizations are investing in chatbot development and deployment. In the 2019 Gartner CIO Survey, CIOs identified chatbots as the main AI-based application used in their enterprises,\footnote{ \url{https://www.gartner.com/smarterwithgartner/chatbots-will-appeal-to-modern-workers/}} with a global market valued in the billions of USD.\footnote{ \url{https://www.mordorintelligence.com/industry-reports/chatbot-market}}

In fact, chatbots are one example of the extent AI technologies are becoming ever more pervasive, both in addressing global challenges, and in the day-to-day routine.
Public administrations too are adopting chatbots for key actions such as helping citizens in requesting services\footnote{\url{https://www.canada.ca/en/employment-social-development/services/my-account/terms-use-chatbot.html}} and providing updates and information, for example, in relation with COVID-19~\cite{chatbotspandemic}.\footnote{\url{https://government.economictimes.indiatimes.com/news/digital-india/covid-19-govt-launches-facebook-and-messenger-chatbot/74843125}}

However, the expansion of intelligent technologies has been met by growing concerns about possible misuses, motivating a need to develop AI systems that are \textit{trustworthy}. On the one hand, governments are pressured for gaining or preserving an edge in intelligent technologies, which make intensive use of large amounts of data. On the other hand, there is an increasing awareness of the need for trustworthy AI systems.\footnote{\url{https://ec.europa.eu/digital-single-market/en/news/ethics-guidelines-trustworthy-ai}}

In the context of information-providing chatbots and assistive dialogue systems, especially in the public sector, we believe that trustworthiness demands transparency, explainability, correctness, and it requires architectural choices that take data access into account from the very beginning.
Arguably, this kind of chatbot should not only use transparent and verifiable methods and be so conceived as to respect relevant data protection regulations, but it should also be able to explain its outputs or recommendations in a manner adapted to the intended (human) user.

We thus propose an architecture for AI dialogue systems where user interaction is carried out \textit{in natural language}, not only for providing information to the user, but also to answer user queries about the \textit{reasons} leading to the system output (explainability). 
The system selects answers based on a \textit{transparent reasoning module}, built on top of a computational argumentation framework with a \textit{rigorous, verifiable semantics} (transparency, auditability).
Additionally, the system has a modular architecture, so as to decouple the natural language interface, where user data is processed, from the reasoning module, where expert knowledge is used to generate outputs (privacy and data governance).

Our work is positioned at the intersection of two areas: computational argumentation and natural language understanding.
While computational argumentation has had significant applications in the context of automated dialogues among software agents, its combination with systems able to interact in natural language in socio-technical systems has been more recent.
The most related proposal in this domain is a recent one by Chalaguine and Hunter~\cite{Hunter}. With respect to such work, our focus is not on persuading the user but on offering correct information. Accordingly, we put greater emphasis on the correctness and justification of system outputs, and on the system's ability to reason with every relevant user input, as opposed to reacting to the last input. Our modular architecture enables a separation between language understanding and argumentative reasoning, which enables significant generality. In particular, our dialogue system architecture can be applied to multiple domains, without requiring any expensive retraining.

In this article we focus on the system's architecture and on the knowledge representation and reasoning module. We start with a brief overview of related approaches (Section~\ref{sec:related}). Next, we give a high-level description of the system architecture (Section~\ref{sec:architecture}) and then zoom in on the argumentation module supporting knowledge representation and reasoning and dialogue strategies (Section~\ref{sec:framework}). To illustrate, we sketch a dialogue between chatbot and human in the context of COVID-19 vaccines (Section~\ref{sec:case}), showing how background knowledge and user data can be formalized and jointly used to provide correct answers, and how the system output can be challenged by the user. Section~\ref{sec:conclusions} concludes.

%% file: s_related.tex
In the field of computational argumentation, significant work has been devoted to defining and reasoning over the argumentation graphs~\cite{Baroni2009,CharwatDGWW15,FazzingaFF19}, leading to several ways of identifying “robust” arguments or sets of arguments~\cite{Dung95,Dung2007}. However,
the practical combination of computational argumentation and dialogue systems based on natural language has not been much explored.
Among the few existing approaches,
Rosenfeld and Kraus~\cite{10.3233/978-1-61499-672-9-320} combine theoretical argumentation with reinforcement learning to develop persuasive agents, while Rach et al.~\cite{10.1007/978-981-13-9443-0_12}  extract a debate's argument structure  and envision the dialogue as a game, structuring the answers as moves along a previously defined scheme. In both cases the agents are limited in their inputs and outputs to sentences ``hard-coded'' in the knowledge base.

An interesting approach in this direction is by 
Chalaguine and Hunter~\cite{Hunter}, who exploit sentence similarity to retrieve an answer from a knowledge base expressed in the form of a graph.
No conversation history is kept, therefore the answers produced by the system do not take into account previous user inputs.
We believe that this approach is inappropriate for complex scenarios where multiple pieces of information must be considered at the same time, since the user would have to include all of them in the same sentence.
Moreover, this approach does not involve reasoning, but relevance-based answer retrieval.
Our approach, instead, aims to output replies `consistent' with \textit{all} the information provided thus far by the user, and that will not be proven wrong later on. In particular, what we do is we \emph{enforce} the condition of \emph{acceptance} of some arguments, by eliciting specific user input. This can be seen as a practical application of the concepts defined by Baumann and Brewka~\cite{BaumannB10}.
In particular, our system relies on
an argumentation module that maintains a history of the concepts expressed by the user and performs reasoning over an argumentation graph to compute the answer. It is therefore possible for the user to consider multiple information at the same time, to  ask for more information if they are needed, and also to provide an explanation for the previous answers.

%% file: s_architecture.tex
Our chatbot architecture consists of two core modules:
the \textit{language module} and the \textit{argumentation module}.
The former provides a natural language interface to the user input, while the latter deals with the problem of computing correct replies to be provided to the user, and it relies on computational argumentation. In this work, we will focus on the argumentation module, leaving the specific implementation of the 
language module for future developments.

We assume the presence of a scenario-specific \emph{knowledge base} (KB) created by experts,
in the form of an argumentation graph (see Section~\ref{sec:argumentation}) with two kinds of nodes.
Nodes are either \emph{status} arguments or  \emph{reply} arguments. The former encode \textit{facts} that correspond to the possible user  sentences. Each status node is linked to one or more \emph{reply} arguments it \emph{supports}\footnote{We point out that our concept of support is a new notion linking status nodes to reply nodes, and its semantics is different from the standard one \cite{CayrolL05a,FazzingaFF18}}, and that
represent replies to the facts stated by the user. Status nodes may also attack other status or reply nodes, typically because the facts they represent are incompatible with one another.
Additionally, a set of natural language sentences is associated with each status node and represents some possible ways a user would express the facts the node encodes. These different representations of facts could be produced by domain experts or crowd-sourced.

\begin{figure}[t]
    \centering
     \includegraphics[width=\columnwidth]{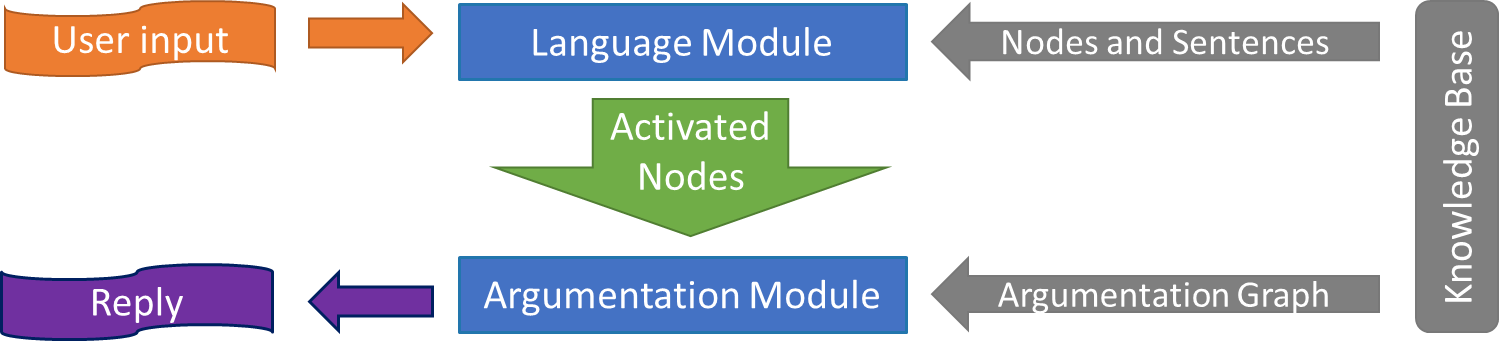}
    \caption{System architecture.}
    \label{fig:architecture}
\end{figure}

The behaviour of the system and the interaction between the modules is illustrated in Figure~\ref{fig:architecture}.
The language module compares each user sentence against the sentences embedded in the KB. In particular, like Chalaguine and Hunter~\cite{Hunter}, we propose to use a sentence similarity measure to identify KB sentences matching the user input. Since each KB sentence is associated with a status node, a list of related status nodes can be computed from the list of sentences in the KB identified by the language module as a match.
%
%
Accordingly, when a user writes a sentence, a set of status nodes $N$ is `activated',  in the sense that they are recognized as matching with the user's input. However, differently from Chalaguine and Hunter~\cite{Hunter}, \emph{all} the status arguments activated during the chat with the user are stored in a set $S$. 

The fundamental principle that characterizes our approach is that a reply $R$ among those supported by $N$ is given to the user only if it is  acceptable w.r.t. $S$. This means that the information given by the user needs to support and defend $R$ from its attacks. 
If there is no acceptable reply with respect to $S$, the chatbot selects anyway a candidate reply $R$, but instead of offering $R$ immediately, it prompts the user in order to acquire new information that could activate new status arguments which, added to $S$, could make $R$ acceptable w.r.t. $S$. This  \textit{elicitation} process 
aims to guarantee that $R$ is not proven wrong in the continuation of the chat. 
In fact, all the information that can be in contrast with $R$ (i.e., that attack $R$) are asked to the user, in order to be sure  to defeat any potential attackers. 

This underlying \emph{strategic reasoning} marks a significant difference from previous approaches. Another distinguishing feature is our system's ability to provide users with online, \emph{on-demand explanations}. In particular, 
besides providing information and getting replies, users can also require an explanation for a given reply $r$. An explanation for $r$  consists of a  sequence of natural language sentences built from $(i)$ descriptions of the status nodes of $S$ supporting $r$ and $ii)$ 
motivations against other possible conflicting replies that the system discarded.


%% file: s_frameworkNEW.tex

The argumentation module is based on a  knowledge base expressed as an argument graph.

\begin{definition}[Argumentation graph]
An argumentation graph is a tuple $\langle A,R,D,T \rangle$, where $A$ and $R$ are the arguments of the graph and are called \emph{status} arguments and \emph{reply} arguments, respectively, $D \subseteq A\times  (A \cup R)$ encodes the attack/defeat relation, and $T \subseteq A\times R$ encodes the support relation.
\end{definition}

Each argument $a$ in $A$ is annotated with a set of natural language sentences, as described in the previous section.
We say that $a$ \emph{attacks} (resp., \emph{supports}) a reply node $r$ iff $(a,r) \in D$ (resp., $(a,r) \in T$). By extension, we say that a set $S$ attacks (resp., supports) $r$, or equivalently that $r$ is attacked by (resp., supported by) $S$, iff there exists an argument $a \in S$ s.t. $a$ attacks (resp., supports) $r$.

The aim of the argumentation module is to identify the reply nodes in response to the user sentences. To this end, 
in addition to the KB, each dialogue session relies on \textit{dynamically acquired knowledge}, expressed as a set of facts or \textit{status arguments} $S$.
The dialogue strategy is to provide the user with a reply that is supported and defended by $S$. However, differently from other proposals, our system does not simply select a consistent reply at each turn. On the contrary, it strategizes in order to provide only robust replies, possibly delaying replies that need further fact-checking. To that end, the two following definitions distinguish between \emph{consistent} and \emph{potentially consistent} reply. The former can be given to the user right away, as it 
can not possibly be proven wrong in the future.\footnote{ The implicit assumption here is that the user does not enter conflicting information, and that the language model correctly interprets the user input. Clearly, if this is not the case, the system's output becomes unreliable. But that wouldn't depend on the underlying reasoning framework. The definition of fall-back strategies able to handle such exceptions would be an important extension to the system.}
The latter, albeit consistent with the current known facts, may still be defeated by future user input, and therefore it should be delayed until a successful elicitation process is completed.

The formal definitions are based on the KB and on a representation of the state of the dialogue consisting of two sets: $S$ and $N$. In particular, $S\subseteq A$ contains the arguments activated during the conversation so far, whereas $N\subseteq S$ contains arguments in support of the system's possible replies to the user.
We recall that  an argument $a$ is \emph{acceptable} w.r.t. a set $S$ iff $S$ defends $a$ from every attack towards $a$.

\begin{definition}[Consistent reply]
\label{def:ar}
Given an argumentation graph $\langle A,R,D,T \rangle$ and two sets $S\subseteq A$ and $N\subseteq S$, a  reply $r\in R$ is \emph{consistent} iff $N$ supports $r$ and $r$  is acceptable w.r.t. $S$.  
\end{definition}

\begin{definition}[Potentially consistent reply]
\label{def:pr}
Given an argumentation graph $\langle A,R,D,T \rangle$ and two sets $S\subseteq A$ and $N\subseteq S$, a  reply $r\in R$ is \emph{potentially consistent} iff $N$ supports $r$, $S$ does not attack $r$ and $r$  is not acceptable w.r.t. $S$.  
\end{definition}

Finally, users can challenge the system output. An \emph{explanation} of a reply $r$ consists of two parts. The first one contains the arguments leading to $r$, i.e., those belonging to a set $S$ that supports $r$. The second one encodes the \textit{why not}s, to explain why the chatbot did not give other replies. 

\begin{definition}[Explanation]
\label{def:expl}
Given an argumentation graph $\langle A,R,D,T \rangle$, a set $S\subseteq A$ and a  reply $r\in R$, an \emph{explanation} for $r$ is a pair $\langle Supp, NotGiven\rangle$, where $Supp$ contains the arguments $a \in S$ s.t. $(a,r) \in T$ and $NotGiven$ is a set of pairs $\langle r',N' \rangle$, where $r'\neq r$, $r'$ is supported by $S$ and $N'\subseteq S$ contains the arguments $b$ attacking $r'$.  
\end{definition}

In the next section we briefly explain how our strategy works to provide the user with consistent replies, by means of
an example in the context of the COVID-19 vaccines.

%% file: s_casestudy.tex
\textit{Disclaimer. The  illustration that follows is based on a (simplistic) representation of the domain knowledge. Its purpose is to show a proof of concept of our approach--not to offer sound advice about vaccines.  We base our example on the content of the AIFA website.\footnote{Italian medicines agency, \url{https://www.aifa.gov.it/en/vaccini-covid-19}.} 
}

\begin{figure}[t]
    \centering
     \includegraphics[width=\columnwidth]{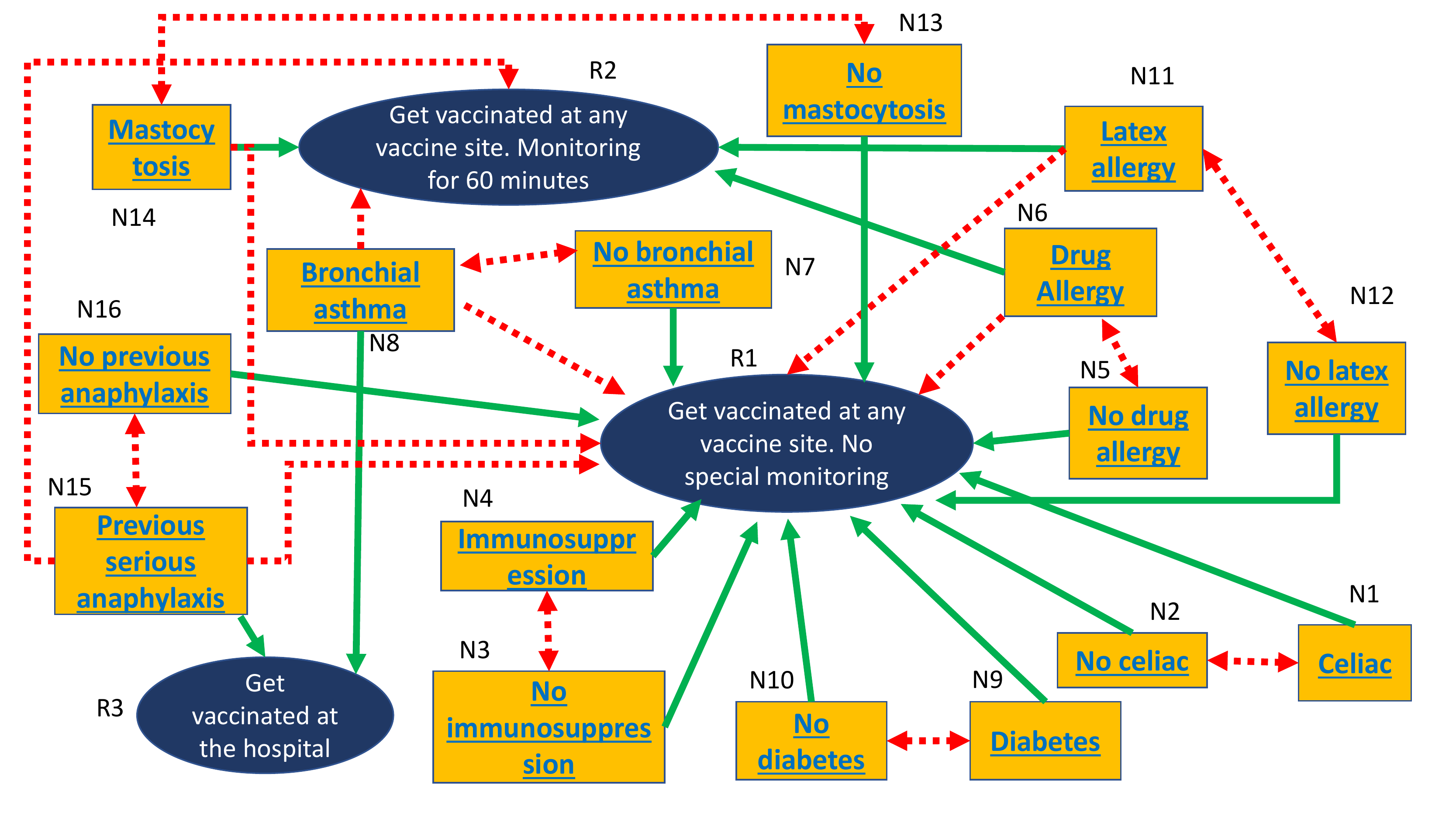}
    \caption{An excerpt of an argumentation graph encoding knowledge about COVID-19 vaccines.}
    \label{fig:KB}
\end{figure}

We consider the context of the vaccines for COVID-19, where we aim to create a dialogue system able to answer  user inquiries about vaccination procedures, vaccine safety, and so on. 
Figure~\ref{fig:KB} shows an excerpt of the argumentation graph encoding the KB, in particular the part related to options for getting vaccinated.

%
%
Yellow rectangles represent  status arguments, blue ovals  reply arguments,  green solid arrows support relations, pointing to the possible replies to  user sentences, and red dotted arrows denote attack relations.
It is worthwhile noticing that the graph contains both the positive and negative version of each status argument. This is a key modeling feature in the context at hand, as it enables the chatbot to properly capture and encode all the information provided by the user about their health conditions.  

Let us consider this example: the user writes ``Hi, I am Morgan and I suffer from latex allergy, can I get vaccinated?” 
The language module processes the user sentence and compares it against all the sentences provided by the knowledge base, resulting in a single positive match with the sentence ``I have latex allergy" associated with node $N_{11}$. 
At this point, the argumentation module deals with the computation of the replies, finding that the only reply supported by  $S=\{N_{11}\}$ is $R_2$ and that it is not a consistent reply, because it is attacked by both $N_{8}$ and $N_{15}$. It is, however, a potentially consistent reply: thus, although we cannot give it yet to the user, what we can do is acquire new information that would make it consistent. 
To make $R_2$ consistent, $S$ must be augmented with both
$N_{7}$ and $N_{16}$. This means that
the user must tell that they do not suffer from bronchial asthma and that they had no previous anaphylaxis.
Then, our strategy is to query the user whether they suffer from bronchial asthma and/or whether they had any previous anaphylaxis. 
Assume at this point that the user replies are $U_1=$\emph{I do not suffer from bronchial asthma} and $U_2=$\emph{I have never had any anaphylaxis}. Then, we can extend $S$ with the new corroborating bits of information, obtaining $S=\{N_{11},N_{7},N_{16}\}$. 
Because
$R_2$  is now a consistent reply, we can return $R_2$ to the user. 

Alternatively, suppose that the user writes that they do suffer from bronchial asthma. In that case,  we would have  $S=\{N_{11},N_{8},N_{16}\}$, hence $R_2$ would not be a consistent reply. 
Accordingly, the only consistent reply that can be given to the user would be
$R_3$.

Finally,
suppose that, upon getting $R_3$ as a reply, the user asks for an explanation.  In that case, 
$\langle Supp, NotGiven\rangle$ is such that $Supp=\{N_8\}$,
and $NotGiven$ consists of the unique pair $\langle R_2,\{N_{8}\} \rangle$, meaning that $R_2$ was not given due to $N_{8}$,
that is, due to the fact that the user suffers from bronchial asthma.

%% file: s_conclusion.tex
We presented a new modular dialogue system architecture based on computational argumentation and language technologies.
In particular, our system exploits both user input and a  knowledge base built by domain experts to perform reasoning in order to compute answers and identify missing bits of information. We illustrated our proposal with an information-seeking scenario, where a user requires information about COVID-19 vaccines.

Our proposal has multiple advantages over previous approaches. With respect to corpus-based dialogue systems, it can use expert knowledge. This is especially important in domains that require trustworthy, correct and explainable solutions. 
Indeed, a remarkable feature of argumentation graphs is their ability to support reasoning over the conflicts between arguments, leading to approving or discarding some responses.
We believe that highlighting the reasons why a response can not be given, along with the facts that rule out other possible responses, is a good way to make the user understand the response and trust the system.
Importantly, the architecture is general-purpose and does not require domain-specific training or reference corpora. 
With respect to prior work on argumentation-based dialogue systems, its major advantage is its ability to reason with multiple elements of user information, in order to provide focused and sound answers, by eventually performing the elicitation of missing data.

In this paper we focused on the argumentation module, leaving the implementation of the language module for future works.  In this regard,  
we plan
to explore the use of recent attention-based neural architectures~\cite{attention}  by representing the user input using BERT-based~\cite{BERT} sentence embeddings~\cite{SBERT} and by comparing them using advanced similarity measures~\cite{cross-lingual}.

Since our proposal is general and not limited to a specific domain, it will be interesting to test our approach on new scenarios and also to consider languages other than English.
Another important aspect we plan to address in the future is the management of conflicting information provided by the user, and the possibility to revise previously submitted information.

%
%